\documentclass[10pt,journal,compsoc]{IEEEtran}
\usepackage{amsmath,amsfonts}
\usepackage{algorithm}
\usepackage{array}
\usepackage[caption=false,font=normalsize,labelfont=sf,textfont=sf]{subfig}
\usepackage{textcomp}
\usepackage{stfloats}
\usepackage{url}
\usepackage{verbatim}
\usepackage{graphicx}
\usepackage{cite}
\usepackage{caption}

\usepackage{changepage}

\usepackage{enumitem}
\usepackage{times}
\usepackage[pagebackref=false,breaklinks=true,colorlinks,bookmarks=false]{hyperref}
\usepackage{cleveref}
\usepackage{multirow}
\usepackage{booktabs}
\usepackage{xspace}

\usepackage{tabularx}
\usepackage{makecell}
\usepackage{algpseudocode}
\usepackage{amssymb}

\makeatletter
\DeclareRobustCommand\onedot{\futurelet\@let@token\@onedot}
\def\@onedot{\ifx\@let@token.\else.\null\fi\xspace}
\def\eg{\emph{e.g}\onedot} 

\def\ie{\emph{i.e}\onedot}

\makeatother

\newcommand{\imagemodel}{{SceneDesigner}\xspace}
\newcommand{\videomodel}{{SceneDesigner++}\xspace}

\newcommand{\imagedata}{{ObjectPose9D}\xspace}
\newcommand{\videodata}{{ObjectPose9D++}\xspace}
\newcommand{\cnocs}{{CNOCS} map\xspace}
\newcommand{\videorope}{VPRoPE\xspace}
\newcommand{\videoropecomplete}{Video Pose RoPE\xspace}
\newcommand{\videoropeone}{VPRoPE-G\xspace}
\newcommand{\videoropetwo}{VPRoPE-L\xspace}

\newcommand{\videobench}{T2IVPoseEval\xspace}

\newcommand{\myparagraph}[1]{%
  \par\noindent\textbf{#1.}\enspace\ignorespaces
}

\crefname{figure}{Fig.}{Figs.}
\Crefname{figure}{Fig.}{Figs.}

\crefname{table}{TABLE}{Tabs.}
\Crefname{table}{TABLE}{Tabs.}

\crefname{equation}{Eq.}{Eqs.}
\Crefname{equation}{Eq.}{Eqs.}

\crefname{section}{Sec.}{Sec.}
\Crefname{section}{Sec.}{Sec.}

\begin{document}

\title{SceneDesigner++: A Unified 9-DoF Pose Control Framework for T2I and T2V Generation}


\author{Xincheng Shuai, Zhenyuan Qin, Henghui Ding, Dacheng Tao,~\IEEEmembership{Fellow,~IEEE}%
\thanks{Project page: \url{https://henghuiding.com/SceneDesigner++/}}
\thanks{Xincheng Shuai, Zhenyuan Qin, and Henghui Ding are with Fudan University, Shanghai, China. E-mail: henghui.ding@gmail.com}
\thanks{Dacheng Tao is with Nanyang Technological University, Singapore.}
}

\markboth{Journal of \LaTeX\ Class Files,~Vol.~14, No.~8, August~2021}%
{Shell \MakeLowercase{\textit{et al.}}: A Sample Article Using IEEEtran.cls for IEEE Journals}


\IEEEtitleabstractindextext{%
\begin{abstract}
Precise 9-DoF pose control is essential for visual content generation, enabling users to specify the location, size, and orientation of objects in 3D space. However, existing approaches typically focus on either text-to-image (T2I) or text-to-video (T2V) generation, with limitations in pose controllability and temporal coherence. In this work, we present \textbf{SceneDesigner++}, a unified framework for 9-DoF pose control across both T2I and T2V generation. To provide comprehensive supervision, we construct \textbf{ObjectPose9D++}, a dataset containing high-quality images and videos with accurate pose annotations. Trained on ObjectPose9D++, SceneDesigner++ combines absolute and relative pose encoding strategies to decouple frame-wise spatial grounding from inter-frame motion modeling. Specifically, absolute pose encoding leverages CNOCS map sequences to provide explicit, frame-wise 9-DoF pose guidance. Meanwhile, our relative pose encoding, termed VPRoPE, incorporates relative pose information into rotary position embeddings to explicitly model inter-frame pose transitions, thereby promoting temporally coherent object motion. A learnable orthogonal transformation is further introduced for each attention head, allowing different heads to capture complementary geometric relationships while preserving input feature norms to promote training stability. We further adopt a two-stage training strategy to learn the absolute and relative pose encoding mechanisms separately, enabling efficient optimization. Benefiting from this lightweight and decoupled design, SceneDesigner++ also supports integration with an independently trained camera-control module, enabling separate or joint control over object and camera motion. Finally, we introduce T2IVPoseEval, a comprehensive benchmark for evaluating diverse object poses and motion patterns in both T2I and T2V tasks. Extensive experiments on this benchmark demonstrate that SceneDesigner++ consistently outperforms same-backbone baselines in pose fidelity and motion quality.
\end{abstract}


}

\maketitle
\IEEEdisplaynontitleabstractindextext

\section{Introduction}\label{introduction}

\begin{figure*}[!t]
    \centering 
\includegraphics[width=\linewidth]{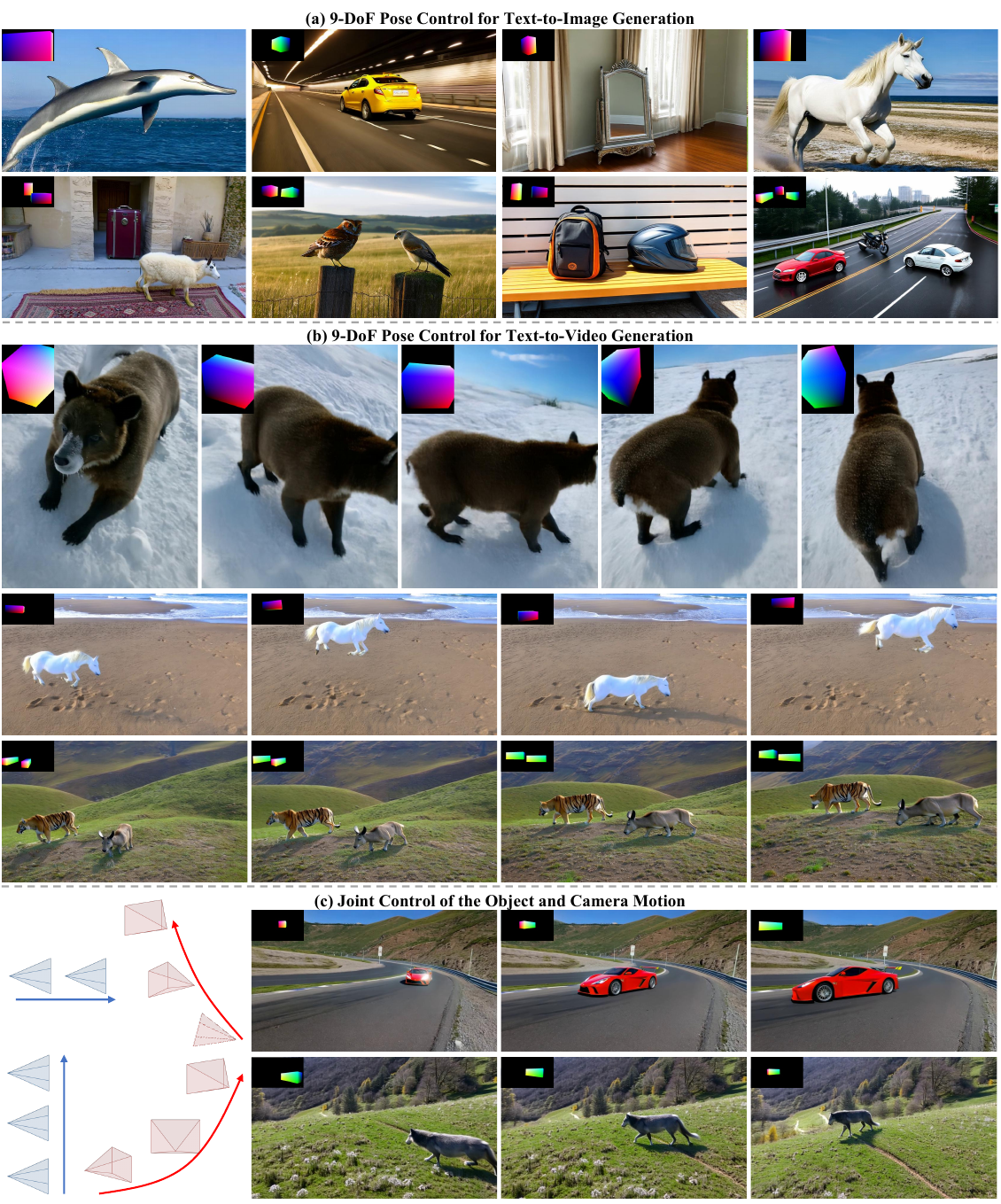}
\vspace{-7.6mm}
    \caption{9D pose control results of the proposed \videomodel for \textbf{(a) text-to-image generation}, \textbf{(b) text-to-video generation}, and \textbf{(c) joint control of object and camera motion}~(The blue and red trajectories represent the camera and object motion, respectively).
    }
    \vspace{-3mm}
\label{fig:teaser}
\end{figure*}

Precise 9D object pose control is essential for applications such as interactive design, film creation, and 3D animation, where users must specify not only scene content but also each object’s pose and motion relative to the camera. To meet this growing demand, an ideal generative model should offer a flexible pose-control interface that unifies both text-to-image (T2I)~\cite{podell2023sdxl,esser2024scaling,dhariwal2021diffusion,flux2024,wu2025qwen,rombach2022high,ho2022classifier,lipman2022flow,peebles2023scalable,nichol2021glide,shuai2026glyphprinter,shuai2024survey,shuai2026psdesigner,liu2026unison} and text-to-video (T2V)~\cite{kong2024hunyuanvideo,guo2023animatediff,blattmann2023stable,hacohen2024ltx,chen2023videocrafter1,chen2024videocrafter2,yang2025cogvideox,zheng2025open,wan2025wan,ho2022imagen} generation. Although many methods~\cite{michel2023object,min2026origen,qiu2024freetraj,ma2024trailblazer,chu2026wan} have been proposed toward this goal, there remains considerable room for improvement in generation quality, control flexibility, and output diversity.

For T2I generation, most early methods~\cite{li2023gligen,xie2023boxdiff} use 2D geometric primitives, \eg, points or bounding boxes, to designate the location and size of an object in the image plane. For example, InstanceDiffusion~\cite{wang2024instancediffusion} projects point coordinates into high-dimensional embeddings and injects them into the diffusion model~\cite{rombach2022high,ho2020denoising} through attention mechanisms. However, such an interface falls short in representing 3D properties, such as depth and orientation, limiting its applications. To address this issue, LOOSECONTROL~\cite{bhat2024loosecontrol} introduces 3D bounding boxes to lift control signals into the 3D space, yet fails to explicitly encode object orientation. Leveraging ground-truth annotations from synthetic datasets, Continuous-3D-Words~\cite{cheng2024learning} learns to generate images with controlled orientation using the pose embedding. Nevertheless, the low-quality synthetic data and the limited diversity of object poses result in poor generalization in real-world scenarios. Moreover, this implicit representation lacks spatial awareness, making it difficult to capture the underlying geometric relationships. To address these issues, our previous method, {\imagemodel}~\cite{qin2026scenedesigner}, introduces {\cnocs}, providing dense guidance by encoding pixel-wise geometric information, thereby enabling faster and more efficient model convergence.

Compared with static images, coherent pose control is more challenging in T2V generation. Most existing methods \cite{wang2024boximator,li2025trackdiffusion,wang2024motionctrl,yin2023dragnuwa,zhang2025tora} simply extend image-space control signals into temporal space. For example, Direct-a-Video~\cite{yang2024direct} and CineMaster \cite{wang2025cinemaster} leverage sequences of 2D/3D bounding boxes to specify object trajectories. However, this straightforward extension inherits the aforementioned limitations of its T2I counterparts \cite{li2023gligen,xie2023boxdiff,bhat2024loosecontrol}. In addition, 2D control signals fail to disentangle object motion from camera dynamics, resulting in ambiguous control. To address this, 3DTrajMaster \cite{fu20253dtrajmaster} embeds per-object 6D poses for more accurate guidance, but is limited by a static camera configuration and the implicit encoding strategy. Free-Form Motion Control~\cite{shuai2025free} employs Pl\"ucker embeddings~\cite{he2024cameractrl} for camera control, while replicating the 3D transformation within the object region to manipulate object pose. Nevertheless, this spatially uniform representation often leads to suboptimal training efficiency.

In summary, achieving accurate and flexible 9D pose control across both T2I and T2V generation remains challenging. This difficulty stems primarily from two challenges. First, most video datasets~\cite{wang2026spatialvid,MOSEv2,MeViS,nan2025openvid,bain2021frozen} lack 9D pose annotations, while data-efficient training paradigms remain underexplored. Although synthetic data~\cite{fu20253dtrajmaster,shuai2025free} provides precise annotations, the synthetic-to-real domain gap often impedes generalization to real-world scenarios. Second, existing pose representations are not well suited to video generation. Although the proposed \cnocs~\cite{qin2026scenedesigner} provides geometrically interpretable, pixel-aligned guidance, efficiently encoding 9D pose sequences remains an open challenge. Our experiments demonstrate that naively conditioning on the sequence of \cnocs leads to suboptimal motion quality. To address these challenges, we develop \textbf{\videomodel}, a unified framework designed for 9D pose control in both T2I and T2V generation, built upon a pretrained T2V foundation model~\cite{wan2025wan}.

\vspace{-2mm}
First, we introduce the \textbf{\videodata} dataset to provide rich supervision for learning generalizable pose control. It extends \imagedata, originally introduced in \imagemodel~\cite{qin2026scenedesigner}, with additional high-quality images and videos accompanied by accurate pose annotations. The video subset comprises samples from domain-specific datasets with ground-truth poses and general-purpose videos annotated using our automated pipeline.

Building upon \videodata, \videomodel combines \textit{absolute and relative pose encoding} to enable precise and coherent pose control. Specifically, the sequence of CNOCS maps provides frame-wise absolute pose guidance and is fused directly with the visual latents at the input layer. This injection mechanism avoids the substantial computational and memory overhead associated with the ControlNet-style branch~\cite{zhang2023adding} used in our previous work, \imagemodel. However, relying solely on absolute pose encoding is insufficient for temporally coherent video generation. Although it specifies the per-frame pose configuration, it fails to explicitly model the relative pose transitions across frames. Therefore, the model has to infer the temporal relationships implicitly, resulting in rigid or unnatural object motion. 

To address this challenge, \videomodel further introduces \textbf{\videorope} (\textbf{\videoropecomplete}), a relative pose encoding scheme that extends RoPE~\cite{heo2024rotary,su2024roformer} by incorporating inter-frame pose relationships into the attention mechanism, thereby enabling the model to capture how object poses evolve over time. Specifically, \videorope integrates pose cues into the rotary matrix of standard RoPE, explicitly conditioning attention scores on the relative poses of objects across frames. This mechanism encourages stronger interactions between geometrically consistent features. To enhance the geometric expressiveness of \videorope, we introduce a learnable orthogonal transformation for each attention head, representing the pose-induced rotations in a head-specific feature basis. In particular, this preserves feature norms to promote training stability while retaining the relative-pose-conditioned attention formulation of \videorope. Meanwhile, head-specific transformations allow each head
to encode pose relationships in a distinct learned subspace,
thereby capturing complementary geometric and motion
variations. Consequently, \videorope improves the temporal coherence and naturalness of object motion with minimal additional parameters and computational overhead.
Benefiting from this design, \videomodel decouples frame-wise spatial grounding from inter-frame motion modeling, enabling accurate pose control while preserving natural dynamics.


Furthermore, we introduce a two-stage training strategy to disentangle the learning of absolute and relative pose encoding.
In the first stage, we optimize only the parameters associated with absolute pose encoding on the T2I data of \videodata, establishing frame-wise 9D pose control capability.
In the second stage, we inject \videorope and optimize only the newly introduced parameters using the T2V data. This training scheme enables the model to focus on learning temporal pose transitions in the second stage. Benefiting from the lightweight and decoupled pose encoding mechanisms, we can additionally incorporate a standalone camera-control module pretrained on camera-annotated data into \videomodel. As shown in \cref{fig:teaser}, the resulting model supports both independent and joint control of object and camera motion. Finally, we introduce \textit{\videobench}, a comprehensive benchmark for evaluating 9D pose accuracy in both image and video generation across diverse motion patterns.

The main contributions of this work are summarized as follows:
\begin{itemize}[leftmargin=1.16em]
    \item We propose \videomodel, a unified framework for 9D pose control in both T2I and T2V generation, providing a flexible interface for specifying the pose trajectories in 3D space. 

   \item The proposed \videodata augments the previous \imagedata dataset with additional images and videos annotated with comprehensive 9D object poses, providing rich supervision for learning accurate and generalizable pose control.

    \item SceneDesigner++ combines two complementary pose encoding strategies: frame-wise CNOCS maps provide absolute pose guidance for accurate spatial grounding, while VPRoPE encodes relative pose transitions for coherent motion modeling. We further introduce learnable head-wise orthogonal transformations to enhance the geometric expressiveness of VPRoPE while preserving feature norms. Together with a two-stage training strategy, these designs improve pose fidelity and temporal motion coherence.

    \item \videomodel supports integration with a standalone camera-control module, enabling independent or joint control of object and camera motion.

    \item We also introduce \videobench, a comprehensive benchmark for evaluating pose accuracy of generated images and videos. Extensive experiments demonstrate that \videomodel consistently outperforms same-backbone baselines across multiple evaluation dimensions.
\end{itemize}

\section{Related Works}
\subsection{Pose Control in Text-to-Image Generation}
Precise pose control facilitates the manipulation of image structure. Early methods~\cite{li2023gligen,xie2023boxdiff,li2025anyi2v}, such as InstanceDiffusion~\cite{wang2024instancediffusion}, process 2D geometric primitives to designate the object arrangement. However, this interface fails to encode 3D properties, such as depth and out-of-plane rotation, limiting the applicability in practical scenarios. Although LOOSECONTROL~\cite{bhat2024loosecontrol} lifts the control signals to 3D space using the 3D bounding box, it still fails to inject orientation information, leaving the underlying geometric ambiguity unresolved. More recently, several 3D-aware methods~\cite{min2026origen,cheng2024learning,qin2026scenedesigner,shuai2025free2} have been proposed to address these limitations. Among training-free approaches, ORIGEN~\cite{min2026origen} builds upon a one-step generation model and optimizes the initial noise using a tailored reinforcement-learning objective, encouraging the poses of generated objects to align with the input conditions. In contrast, training-based methods, such as Continuous-3D-Words~\cite{cheng2024learning} and our previous work \imagemodel~\cite{qin2026scenedesigner}, utilize the pose annotations from synthetic data or real-world images, enabling more flexible control. Among them, Continuous-3D-Words utilizes a spatially unaware pose embedding, which is limited by poor geometric interpretation, leading to inefficient learning. To address this issue, \imagemodel introduces \cnocs, a pixel-wise representation that enables faster model convergence. Moreover, by training on real-world data, \imagemodel improves generation quality, pose diversity, and control flexibility over previous methods.

\subsection{Motion Control in Text-to-Video Generation}
Effectively manipulating video dynamics has emerged as an important task in video generation. One line of research focuses on object motion control~\cite{WBench}. Early approaches~\cite{li2025trackdiffusion,zhang2025tora,qiu2024freetraj,ma2024trailblazer,chu2026wan}, such as DragNUWA~\cite{yin2023dragnuwa} and Boximator~\cite{wang2024boximator}, take trajectories of 2D geometric primitives as input to manipulate motion fields in image space. For example, Boximator and Direct-a-Video~\cite{yang2024direct} employ the 2D bounding box sequence to control the per-object motion trajectory. Moreover, 3DTrajMaster~\cite{fu20253dtrajmaster} was learned on synthetic data to handle 6D pose trajectories. However, its generalization to real-world scenarios is hindered by the limited diversity of motion patterns and the lack of realism in synthetic data.

On the other hand, a group of methods~\cite{bahmani2025vd3d,bahmani2025ac3d} condition video generation on camera movement to control global scene dynamics. CameraCtrl~\cite{he2024cameractrl} converts intrinsic and extrinsic parameters into the pixel-aligned Pl\"ucker embedding and injects it into the main backbone through a camera encoder. This representation has subsequently been widely adopted in the following studies~\cite{shuai2025free,bahmani2025ac3d}. Inspired by related advances in image generation~\cite{safin2023repast,miyato2024gta,kong2024eschernet,li2026cameras}, another line of work~\cite{zhang2026unified,yin2026scope,li2026rerope} incorporates camera parameters directly into positional encoding. By embedding camera geometry into visual tokens, these methods establish geometry-aware correspondences across viewpoints, leading to improved control over camera motion. Among them, UCPE~\cite{zhang2026unified} further generalizes camera encoding beyond a fixed pinhole-camera model, allowing the model to accommodate more diverse camera configurations within a unified formulation.

Several methods have sought to jointly control object and camera motion. Early methods such as MotionCtrl~\cite{wang2024motionctrl} parameterize object motion in 2D image space, resulting in incomplete disentanglement between object and camera dynamics and frequently producing physically implausible motion. In contrast, CineMaster~\cite{wang2025cinemaster} jointly conditions generation on sequences of 3D bounding boxes and camera poses, partially alleviating this ambiguity. Free-Form Motion Control~\cite{shuai2025free} further injects orientation information for more flexible control. Nevertheless, its reliance on synthetic training data limits its performance in real-world scenarios. 

\section{Preliminaries}\label{sec:preliminary}
\subsection{Definitions of 9-DoF Pose Control and CNOCS Map}
\label{sec:9d_pose_task}

For the T2I generation task, given the prompt $c_p$ and 9D poses of $M$ objects $\left\{\mathcal{P}_i\right\}_{i=1}^{M}$, the model is required to generate the image aligned with both text and pose descriptions. Specifically, the 9D pose $\mathcal{P}$ is defined as $\mathcal{P}=(\mathbf{l},\mathbf{s},\mathbf{r})$, where $\mathbf{l}$, $\mathbf{s}$, and $\mathbf{r}$ denote the object's 3D location, size, and orientation, respectively. The definition can be easily extended to T2V generation, where the pose condition becomes its temporal extension $\left\{\left\{\mathcal{P}_i^f\right\}_{i=1}^{M}\right\}_{f=1}^{F}$, where $F$ is the total number of the frames.

In our previous conference paper, \imagemodel~\cite{qin2026scenedesigner} was proposed to achieve 9D pose control for T2I generation. Specifically, it extends the Normalized Object Coordinate System (NOCS)~\cite{wang2019normalized} to a more generalized representation, \cnocs, enabling the pretrained T2I models~\cite{esser2024scaling} to accomplish explicit pose control. Concretely, the ordinary NOCS establishes dense correspondences between image pixels and normalized object coordinates. To achieve this, it constructs a shared canonical space for each object category, while encoding each pixel's normalized 3D coordinates as RGB values. Despite its effectiveness, NOCS requires the exact object geometry, \eg, CAD model, limiting its application. Therefore, \cnocs leverages the 3D cuboid as the shape proxy to alleviate the constraint. Compared with the spatially unaware pose embedding adopted in the previous literature~\cite{fu20253dtrajmaster,cheng2024learning}, this pixel-aligned representation exhibits strong geometric interpretation, while providing dense guidance for efficient supervision. We follow the similar formalization established in the conference version and express the construction process of \cnocs as:
\begin{equation}
\begin{aligned}
\label{eq:cnocs}
x^c_{u,v},y^c_{u,v},z^c_{u,v},\mathrm{o}_i
&=\operatorname{Intersect}\!\left(u,v,\{\mathrm{bbox}_i\}_{i=1}^{M}\right),\\
\left[x^o_{u,v},y^o_{u,v},z^o_{u,v},1\right]^\top
&=\mathbf{T}^{\mathrm{o}_i}_{\mathrm{c2o}}\!\left[x^c_{u,v},y^c_{u,v},z^c_{u,v},1\right]^T,\\
x^{no}_{u,v},y^{no}_{u,v},z^{no}_{u,v}
&=\operatorname{normalize}\!\left(x^o_{u,v},y^o_{u,v},z^o_{u,v},\mathrm{bbox}_{\mathrm{o}_i}\right),\\
\operatorname{CNOCS}[u,v]
&=\operatorname{render}\!\left(x^{no}_{u,v},y^{no}_{u,v},z^{no}_{u,v}\right).
\end{aligned}
\end{equation}
For a pixel indexed by $(u,v)$, a camera ray is intersected with the set of oriented object cuboids $\{\mathrm{bbox}_i\}_{i=1}^{M}$ using the ``$\operatorname{Intersect}$'' operation, which returns the visible surface point $(x^c_{u,v},y^c_{u,v},z^c_{u,v})$ and its object index $\mathrm{o}_i$. Then, the homogeneous matrix $\mathbf{T}^{\mathrm{o}_i}_{\mathrm{c2o}} \in \mathbb{R}^{4\times 4}$ transforms this point from the camera space to the corresponding object space. The coordinate $\left(x^{o}_{u,v},y^{o}_{u,v},z^{o}_{u,v}\right)$ is then normalized by the lengths of the cuboid sides, obtaining $\left(x^{no}_{u,v},y^{no}_{u,v},z^{no}_{u,v}\right)$. Finally, the pixel value of the \cnocs is calculated by applying the rendering function $\operatorname{render}: \mathbb{R}^{3}\rightarrow\mathbb{R}^{3}$, which is selected as the identity function in \imagemodel, \ie, $\operatorname{CNOCS}[u,v]=\left(x^{no}_{u,v},y^{no}_{u,v},z^{no}_{u,v}\right)$.

\subsection{\imagemodel}
\imagemodel~\cite{qin2026scenedesigner} is our previous T2I framework designed for multi-object 9D pose control. It introduces a ControlNet-like branch~\cite{zhang2023adding} that processes the \cnocs and injects the resulting geometric features into the frozen generation backbone. For training, \imagemodel constructed the \imagedata dataset by combining samples from OmniNOCS~\cite{krishnan2024omninocs} and MS-COCO~\cite{lin2014microsoft} with comprehensive 9D pose annotations, followed by manual filtering and refinement. The model was trained in two stages. The first stage optimizes the introduced control branch on ObjectPose9D using the flow matching objective~\cite{esser2024scaling,lipman2022flow}. The second stage performs reinforcement learning to improve low-frequency pose generation, \eg the back views. Specifically, the designed reward evaluates the fidelity of the generated object to the target location, size, and orientation, while a prior preservation loss maintains generation quality and prevents reward hacking. During inference, \imagemodel further employs ``Disentangled Object Sampling'', which separately predicts global and object-specific velocity fields and combines them using object masks. This strategy associates each object with its designated pose while alleviating attribute leakage in multi-object generation.

\subsection{Video Attention and Rotary Position Embedding}
\label{sec:video_attention_rope}
In this section, we revisit the attention mechanism used in DiT-based T2V models~\cite{yang2025cogvideox,wan2025wan,hacohen2024ltx}. Given the noisy video latent ${\mathbf{z}} \in \mathbb{R}^{F\times H \times W \times C}$, existing models first flatten it into a sequence of $N=FHW$ visual tokens ${\hat{\mathbf{z}}}  \in \mathbb{R}^{N\times C}$, where each token is associated with a coordinate $\mathbf{p}=(t,h,w)$ in time-height-width dimensions. Then, the self-attention projects $\hat{\mathbf{z}}$ into the query, key, and value $\mathbf{Q},\mathbf{K},\mathbf{V}\in\mathbb{R}^{N\times d}$, where $d$ is the dimension of the hidden state:
\begin{equation}
    \mathbf{Q}={\hat{\mathbf{z}}}\mathbf{W}^{Q},\qquad
    \mathbf{K}={\hat{\mathbf{z}}}\mathbf{W}^{K},\qquad
    \mathbf{V}={\hat{\mathbf{z}}}\mathbf{W}^{V},
\end{equation}
where $\mathbf{W}^{Q}$, $\mathbf{W}^{K}$ and $\mathbf{W}^{V}$ are learnable matrices. Then, the attention output is computed as:
\begin{equation}\label{eq:attention}
\begin{aligned}
    &\operatorname{Attn}(\mathbf{Q},\mathbf{K},\mathbf{V})
    =
    \mathbf{A}\mathbf{V},\\
    &\mathbf{A}
    =
    \operatorname{Softmax}
    \left(
        \frac{f\left(\mathbf{Q},\mathbf{P}\right)^\top f\left(\mathbf{K},\mathbf{P}\right)}{\sqrt{d}}
    \right),
\end{aligned}
\end{equation}
where $\mathbf{A}\in\mathbb{R}^{N\times N}$ is the attention weight matrix, $\mathbf{P}\in\mathbb{R}^{N\times 3}$ concatenates the token coordinate $\mathbf{p}$ in order, and $f$ denotes the positional encoding function. In particular, $\mathbf{A}_{mn}$ represents the normalized attention score, quantifying the contribution of the $n$-th token to the updated value of the $m$-th token. For most generative models~\cite{bai2025qwen2,wu2025qwen,yang2025qwen3}, Rotary Position Embedding (RoPE)~\cite{su2024roformer} is widely used as the $f$ operation, where the attention score is calculated conditioned on the relative positional offset. 
Concretely, the one-dimensional RoPE applies a transformation $\mathbf{R}_{d\times d}$ to the vector input $\mathbf{x}\in\mathbb{R}^{1\times d}$ according to its position $m$:
\begin{equation}
\begin{aligned}
\label{eq:rope_1d}
&f_{\text{RoPE-1D}}(\mathbf{x},m)
=\mathbf{R}_{d\times d}(m)\mathbf{x}^\top\\
&=\operatorname{BD}_{i=0}^{d/2-1}\!\left(
\begin{bmatrix}
\cos(m\theta_i) & -\sin(m\theta_i)\\
\sin(m\theta_i) & \cos(m\theta_i)
\end{bmatrix}
\right)\mathbf{x}^\top,
\end{aligned}
\end{equation}
where ``$\operatorname{BD}$'' constructs a block-diagonal matrix and $\theta_i=\theta_\text{base}^{-2i/d}$ (typically $\theta_\text{base}= 10000$) determines the channel-wise frequency. Benefiting from this, the unnormalized attention score between the rotated $m$-indexed query and $n$-indexed key can be calculated as:
\begin{equation}
\begin{aligned}
\label{eq:f_dot}
    &f_{\text{RoPE-1D}}(\mathbf{Q}_m,m)^\top f_{\text{RoPE-1D}}(\mathbf{K}_n,n) \\
    &=
    \mathbf{Q}_m
    {\mathbf{R}}_{d \times d}^{\top}(m)
    {\mathbf{R}_{d \times d}}(n)
    \mathbf{K}_n^{\top} \\
    &=
    \mathbf{Q}_m
    {\mathbf{R}_{d \times d}}(n-m)
    \mathbf{K}_n^{\top},
\end{aligned}
\end{equation}
where the subscript of $\mathbf{Q}, \mathbf{K}$ denotes the corresponding matrix row. Thus, the resulting attention score depends on the relative token positions. Furthermore, \cref{eq:rope_1d} can be extended to its 3D-variant for video input:
\begin{equation}
\begin{aligned}
\label{eq:rope_3d}
&f_{\text{RoPE-3D}}(\mathbf{x},\mathbf{p}=(t,h,w))=\bar{\mathbf{R}}_{d\times d}(\mathbf{p})\mathbf{x}^\top \\
&=
    \begin{bmatrix}
        \mathbf{R}_{d_t\times d_t}(t) & \mathbf{0} & \mathbf{0} \\
        \mathbf{0} & \mathbf{R}_{d_h\times d_h}(h) & \mathbf{0} \\
        \mathbf{0} & \mathbf{0} & \mathbf{R}_{d_w\times d_w}(w)
    \end{bmatrix} \mathbf{x}^\top \\
&=[f_{\text{RoPE-1D}}(\mathbf{x}[0:d_t],t)^\top,f_{\text{RoPE-1D}}(\mathbf{x}[d_t:d_t+d_h],h)^\top,\\
&f_{\text{RoPE-1D}}(\mathbf{x}[d_t+d_h:d],w)^\top]^\top,
\end{aligned}
\end{equation}
where $d=d_t+d_h+d_w$ and $[:]$ denotes the slice operation. Thus, the attention score is conditioned on the relative temporal and spatial positions between two video tokens. 
We can also express the matrix version of 3D-RoPE as:
\begin{equation}
\begin{aligned}
\label{eq:rope_3d_matrix}
f_{\text{RoPE-3D}}(\mathbf{X},\mathbf{P})
=[&f_{\text{RoPE-3D}}\left(\mathbf{X}_1,\mathbf{P}_1\right)^\top;f_{\text{RoPE-3D}}\left(\mathbf{X}_2,\mathbf{P}_2\right)^\top;\\
&...;f_{\text{RoPE-3D}}\left(\mathbf{X}_{N},\mathbf{P}_N\right)^\top],
\end{aligned}
\end{equation}
where $\mathbf
{X} \in \mathbb{R}^{N\times d}$ denotes the matrix input (\eg, $\mathbf{Q}$ and $\mathbf{K}$) and $;$ splits the matrix rows.

\begin{table*}[t]
    \centering
    \caption{
    Statistics of the additional data incorporated into \videodata beyond \imagedata. 
    }
    \label{tab:dataset_statistic}
    \setlength{\tabcolsep}{8pt}
    \resizebox{\textwidth}{!}{
    \begin{tabular}{lccccc}
        \toprule
        Modality & Source & Data Type
        & \# Samples & Resolution~($F \times H\times W$) & Pose Annotation \\
        \midrule
        \multirow{2}{*}{Image}
        & Internal Data
        & Real
        & 6,561 & $1 \times 480 \times 832$ 
        & SAM 3 + SAM 3D \\
        & Synthetic Simulation
        & Synthetic
        & 10,000 & $1 \times 480 \times 832$  
        & Ground-Truth Engine State \\
        \midrule
        \multirow{4}{*}{Video}
        & Internal Data
        & Real
        & 5,943 & $81\times480 \times 832$  
        & SAM 3 + SAM 3D  \\
        & Objectron~\cite{ahmadyan2021objectron}
        & Real
        & 10,469  & $81\times 832 \times 640$  
        & 3D Transformation+3D Bounding Box \\
        & PFERD~\cite{li2024poses}
        & Real
        & 6,590  & $81\times480 \times 832$  
        & 3D Transformation+hSMAL shape \\
        & CARLA~\cite{dosovitskiy2017carla}
        & Synthetic
        & 8,538 & $81\times480 \times 832$ 
        & Ground-Truth Engine State \\
        \bottomrule
    \end{tabular}}
\end{table*}

\section{Method}\label{sec:method}
\subsection{\videodata}\label{sec:method_dataset}
We first introduce \textbf{\videodata}, which extends our previous \imagedata~\cite{qin2026scenedesigner} dataset with high-quality image and video data accompanied by accurate pose annotations, enabling unified training for both T2I and T2V generation. 

\myparagraph{Data Sources}~Since \imagedata contains images of relatively limited quality and partially noisy annotations, we first augment it with high-quality internal images and synthetically rendered images. For the synthetic subset, we follow a similar construction pipeline established in prior works~\cite{shuai2025free2}. In particular, we collect a large number of 3D assets from publicly available websites and discard samples whose orientations are inherently ambiguous. Using the Cycles rendering engine in Blender~\cite{blender}, we procedurally build scenes containing one to three objects and randomize their spatial arrangement, illumination, backgrounds, and camera viewpoints to produce the final images. 
On the other hand, the video subset of \videodata is primarily built upon the following sources:
\begin{itemize}[leftmargin=0.96em]
    \item We leverage the internal high-quality videos, covering diverse motion patterns and object categories.  
    \item Objectron~\cite{ahmadyan2021objectron} contains object-centric videos captured in real-world indoor and outdoor environments. Each sequence records a moving camera capturing the static object, covering nine common categories. The dataset provides oriented 3D bounding boxes and 3D transformations for each frame.
    \item PFERD~\cite{li2024poses} comprises horse motion sequences captured from multiple synchronized and calibrated camera views, annotated with per-frame hSMAL~\cite{li2021hsmal} shape and 3D transformation. 
    \item We also construct a synthetic dataset using the CARLA simulator~\cite{dosovitskiy2017carla}, which contains controllable urban traffic sequences involving moving vehicles, pedestrians, cyclists, and static roadside objects under diverse road layouts and weather settings.
\end{itemize}
Furthermore, \cref{tab:dataset_statistic} summarizes the additional data incorporated into \videodata beyond \imagedata. In total, we add $16,561$ images and $31,540$ video clips from both synthetic and real-world sources.

\begin{figure*}[t]
    \centering
    \includegraphics[width=\textwidth]{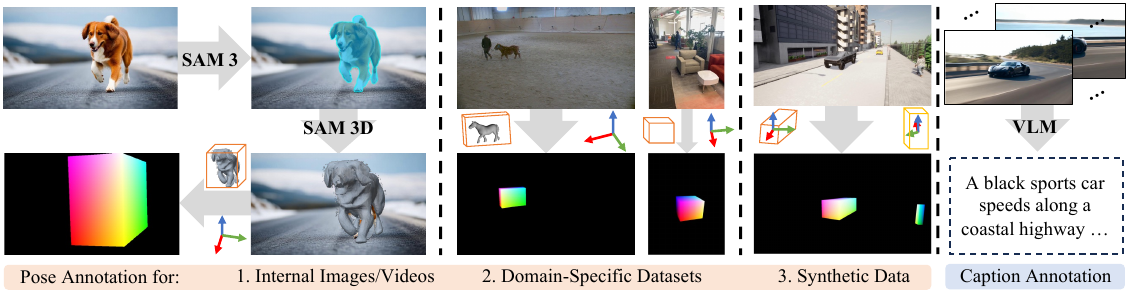}
    \caption{Overview of our annotation pipeline. For constructing \cnocs, the internal images and videos are annotated using SAM 3 and SAM 3D, while domain-specific datasets, including Objectron and PFERD, use their provided pose annotations. Synthetic data are annotated with ground-truth 9D poses recorded by Blender and CARLA engines. 
    Finally, a VLM is employed to generate captions for the images and videos.}
    \label{fig:annotation_pipeline}
\end{figure*}

\myparagraph{Annotation Pipeline}~An overview of our annotation pipeline is provided in \cref{fig:annotation_pipeline}. Depending on the data source, object poses are obtained from model-based estimation, intrinsic annotations, or ground-truth states from rendering engines. Following \cref{eq:cnocs}, we can easily obtain \cnocs by constructing 3D bounding boxes and the camera-to-object transformation $\mathbf{T}_\text{c2o}$. 
\begin{itemize}[leftmargin=0.96em]
    \item For internal data, we first use SAM 3~\cite{carion2026sam} to extract object masks. The estimated masks are then provided to SAM 3D~\cite{chen2026sam} to infer the object meshes and $\mathbf{T}_\text{c2o}$. Specifically, we can leverage the mesh to derive the 3D bounding box, while $\mathbf{T}_\text{c2o}$ can be used to further convert it into \cnocs. For videos, we discard samples with incomplete or discontinuous estimations. Additionally, we also filter out incorrectly annotated samples.

    \item For domain-specific datasets~\cite{ahmadyan2021objectron,li2024poses}, we leverage the intrinsic annotations in Objectron~\cite{ahmadyan2021objectron} and PFERD~\cite{li2024poses} to derive the \cnocs. In particular, the 3D bounding box and the transformation $\mathbf{T}_\text{c2o}$ are already recorded in Objectron. For PFERD, we utilize the hSMAL shape to render the meshes, which can be used to construct the object cuboid.

    \item For synthetic data, we use the rendering engine, \ie, Blender and the CARLA simulator, to programmatically record the 9D pose of each object, which can be converted into \cnocs.
\end{itemize}
Finally, we employ a powerful VLM~\cite{bai2025qwen3} to generate captions for images and videos. To improve efficiency, we uniformly sample ten frames from each video for captioning.

\begin{table*}[t]
    \centering
   \caption{
Comparison of representative methods for pose-controllable image and video generation.
The ``{Obj./Cam. Decoupled}'' column indicates whether the object and camera motion can be controlled independently without disentanglement.
Among these methods, \videomodel is the only unified T2I\&T2V framework that supports generalizable 9D pose control while explicitly decoupling object and camera motion.
}
    \label{tab:method_comparison}
    \setlength{\tabcolsep}{1pt}
    \resizebox{\textwidth}{!}{
    \begin{tabular}{lccccc}
        \toprule
        Method
        & Pose Representation 
        & \makecell{Controllability}
        & Task
        & \makecell{Obj./Cam. Decoupled}
        & Training Data \\
        \midrule
        InstanceDiffusion~\cite{wang2024instancediffusion}
        & 2D Geometric Primitives
        & 2D Location\&Size
        & T2I
        & -
        & \textbf{Real-World Data}  \\
        LOOSECONTROL~\cite{bhat2024loosecontrol}
        & 3D Bounding Box
        & 3D Location\&Size
        & T2I
        & -
        & \textbf{Real-World Data} \\
        Continuous-3D-Words~\cite{cheng2024learning}
        & Implicit Pose Embedding
        & 1D Orientation
        & T2I
        & -
        & Synthetic Data \\
        ORIGEN~\cite{min2026origen}
        & Euler Angles
        & 3D Orientation
        & T2I
        & -
        & -  \\
        \imagemodel~\cite{qin2026scenedesigner}
        & \textbf{CNOCS Map}
        & \textbf{9D Pose}
        & T2I
        & -
        & \textbf{Real-World Data}  \\
        \midrule
        Direct-a-Video~\cite{yang2024direct}
        & 2D Bounding Box Sequence
        & 2D Location\&Size
        & T2V
        & No
        & -  \\
        Boximator~\cite{wang2024boximator}
        & 2D Bounding Box Sequence
        & 2D Location\&Size
        & T2V
        & No
        & \textbf{Real-World Data} \\
        MotionCtrl~\cite{wang2024motionctrl}
        & 2D Point Trajectory
        & 2D Location
        & T2V
        & No
        & \textbf{Real-World Data} \\
        CineMaster~\cite{wang2025cinemaster}
        & 3D Bounding Box Sequence
        & 3D Location\&Size
        & T2V
        & \textbf{Yes}
        & \textbf{Real-World Data} \\
        3DTrajMaster~\cite{fu20253dtrajmaster}
        & Implicit Pose Embedding
        & 3D Location\&Orientation
        & T2V
        & No
        & Synthetic Data  \\
        Free-Form Motion Control~\cite{shuai2025free}
        & 7D Spatial Map
        & 3D Location\&Orientation+1D Size
        & T2V
        & \textbf{Yes}
        & Synthetic Data  \\
        \midrule
        \textbf{\videomodel}~(\textbf{Ours})
        & \textbf{\cnocs Sequence+\videorope}
        & \textbf{9D Pose}
        & \textbf{T2I+T2V}
        & \textbf{Yes}
        & \textbf{Real-World \& Synthetic Data}  \\
        \bottomrule
    \end{tabular}}
\end{table*}

\subsection{\videomodel}\label{sec:method_method}
We propose \videomodel in this work, a unified framework for 9D pose control across both T2I and T2V generation. \videomodel is built upon a pretrained T2V model~\cite{wan2025wan}, while conditioning on object poses through two complementary mechanisms. \textbf{(1) Absolute pose encoding.} We employ a sequence of \cnocs to provide dense, frame-wise pose conditions. Each video frame is thus conditioned on pixel-aligned geometric cues, promoting accurate object-pose control for each video frame. \textbf{(2) Relative pose encoding.} The above absolute mechanism specifies the pose of each frame independently but fails to explicitly capture pose transitions or correspondences across frames, making it insufficient for coherent motion modeling. To address this limitation, \videomodel introduces \videorope, which refines the relative positional encoding to condition attention on inter-frame pose relationships, thereby reducing rigid and unnatural motion. We further devise a two-stage training scheme that decouples the optimization of the parameters associated with these two encoding mechanisms. \cref{tab:method_comparison} compares \videomodel with some representative pose-controllable generation methods from several perspectives. As shown in the table, only our method supports comprehensive 9D pose control across both T2I and T2V tasks. 
Benefiting from high-quality training data and a disentangled encoding strategy, our method generalizes effectively to the real-world domain while enabling joint control of object and camera motion. The remainder of this section introduces \videomodel in detail.

\subsubsection{Stage I: Learning Absolute Pose Encoding}
In this stage, we aim to equip the pretrained T2V model with fundamental 9D pose control capabilities. To this end, we leverage the \cnocs sequence to provide dense, frame-wise guidance through a geometrically interpretable and pixel-aligned representation. As formalized in ~\cref{eq:cnocs_spatial}, we replace the computationally intensive ControlNet-style branch~\cite{zhang2023adding} adopted in \imagemodel with a lightweight input-level conditioning mechanism.
\begin{equation}
\begin{aligned}
\label{eq:cnocs_spatial}
\hat{\mathbf{z}}^\prime=\Phi_\theta^\text{proj}(\text{Concat}(\hat{\mathbf{z}},\Phi_\theta^\text{cnocs}(\text{CNOCS}))),
\end{aligned}
\end{equation}
where $\Phi_\theta^\text{cnocs}$ projects the raw \cnocs into the same space as the noisy visual latent $\hat{\mathbf{z}}$. Then, the encoded \cnocs is concatenated with $\hat{\mathbf{z}}$ along the channel dimension, and $\Phi_\theta^\text{proj}$ subsequently transforms the fused features into the input space of the pretrained backbone. The resulting representation $\hat{\mathbf{z}}^\prime$ is processed by the remaining modules of the main network. 

Since \cref{eq:cnocs_spatial} independently specifies the 9D pose configuration for each frame, we refer to it as \textit{absolute pose encoding}. To avoid inter-frame interactions and allow the model to focus on learning the spatial correspondence between pixel-aligned \cnocs and object appearance under the target pose, we only optimize the introduced parameters of $\Phi_\theta^\text{cnocs},\Phi_\theta^\text{proj}$ on T2I data. In particular, both the image and video subsets from \videodata are utilized for training: for a video sample, an individual frame and its corresponding \cnocs are randomly selected to form a text-image-CNOCS triplet. For training, we leverage the following flow matching objective~\cite{esser2024scaling}:
\begin{equation}
\begin{aligned}
\label{eq:loss_flow}
&\mathcal{L}_{\mathrm{flow,T2I}}=
\mathbb{E}_{\substack{t,(\mathbf{z}_0,c_p,\text{CNOCS})\sim\mathcal{D}_\text{T2I},\epsilon\sim\mathcal{N}(\mathbf{0},\mathbf{I})}}\\
&\left[
\left\|v_\theta\!\left(\mathbf{z}_t,t,c_p,\text{CNOCS}\right)
-(\epsilon-\mathbf{z}_0)\right\|_2^2
\right],
\end{aligned}
\end{equation}
where a clean image latent $\mathbf{z}_0$ and the text prompt $c_p$ are sampled from the constructed T2I data $D_\text{T2I}$, and $\mathbf{z}_t$ is calculated by $\mathbf{z}_t=t\epsilon+(1-t)\mathbf{z}_0$. $v_\theta$ predicts the velocity field. 
We also employ an auxiliary regional loss to enhance the loss weights of tokens associated with pose control:
\begin{equation}
\begin{aligned}
\label{eq:loss_total_stage1}
\mathcal{L}_{\text{Stage-I}}&=\mathcal{L}_{\mathrm{flow,T2I}}+\gamma \mathcal{L}_{\mathrm{region}},\\
\mathcal{L}_{\mathrm{region}}&=\mathbb{E}\left[
\left\|\mathbf{M}_\text{obj} v_\theta\!\left(\mathbf{z}_t,t,c_p,\text{CNOCS}\right)
-\mathbf{M}_\text{obj}(\epsilon-\mathbf{z}_0)\right\|_2^2
\right],
\end{aligned}
\end{equation}
where $\gamma$ is the weight coefficient and $\mathbf{M}_\text{obj}$ denotes the object mask derived from \cnocs.

\begin{figure*}[!t]
    \centering
\includegraphics[width=\linewidth]{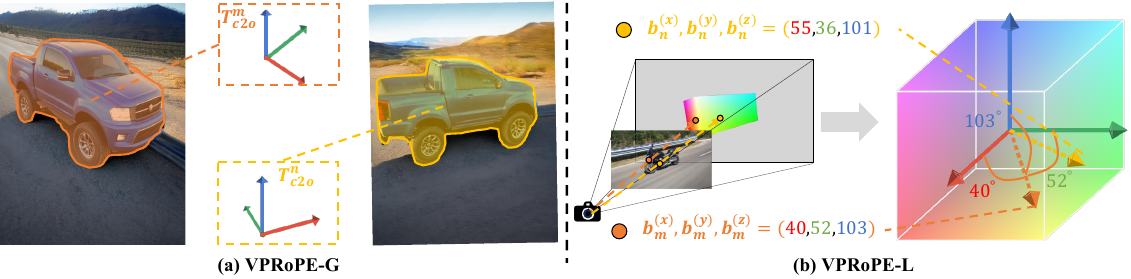}
    \caption{The figure illustrates the two proposed formulations of VPRoPE, where different colors differentiate the query and key tokens. \textbf{(a) \videoropeone} performs object-level encoding. Under this formulation, the RoPE matrices of tokens within the same object region are determined by the camera-to-object transformation $\mathbf{T}_{\mathrm{c2o}}$ of the corresponding frame, establishing relative pose relationships across frames. In contrast, \textbf{(b) \videoropetwo} provides a finer-grained encoding scheme. Specifically, it uses the per-pixel normalized coordinates recorded in \cnocs and derives positional indices from the corresponding direction angles (bin width is set to 1 for this example). This formulation preserves the multi-frequency structure of standard RoPE while establishing pixel-level correspondences, thereby offering better representational capacity.}
    \label{fig:videorope}
\end{figure*}

\subsubsection{Stage II: Learning Relative Pose Encoding}
After Stage I, the model has acquired accurate frame-level pose control through absolute encoding. The remaining challenge is to extend this capability to temporally coherent 9D pose control. A direct solution is to further fine-tune the model on the T2V data of \videodata, conditioning on the sequence of \cnocs. However, our experiments show that this approach tends to produce rigid and unnatural object dynamics. 
We attribute this limitation to the absence of explicit geometric cues for inter-frame pose transitions. 
To address this issue, we introduce \textbf{\videorope}~(\textbf{\videoropecomplete}) as a \textit{relative pose encoding} that complements the absolute mechanism. Specifically, it modulates the positional encoding such that attention scores are explicitly conditioned on relative object poses across frames, thereby promoting temporally coherent motion. Below, we discuss the two formulations of \videorope.

\myparagraph{\videoropeone}
We first introduce a global variant, \videorope-\underline{G}, encoding object-level pose transitions across frames. In particular, it leverages the \underline{g}lobal camera-to-object transformation $\mathbf{T}_\text{c2o}$ to reformulate the RoPE matrix. Under this formulation, we construct the transformation $\bar{\mathbf{R}}_{d\times d}^{\text{global}}$ by replicating $\mathbf{T}_\text{c2o}$ along the block diagonal:
\begin{equation}
\begin{aligned}
\label{eq:f_vrope_v1}
f_{\text{global}}(\mathbf{x},\mathbf{T}_\text{c2o})
&=\bar{\mathbf{R}}_{d\times d}^{\text{global}}(\mathbf{T}_\text{c2o})\mathbf{x}^\top=(\mathbf{I}_{d/4 \times d/4}\otimes\mathbf{T}_\text{c2o})\mathbf{x}^\top,
\end{aligned}
\end{equation}
where $\otimes$ denotes the Kronecker product. Then, we can calculate the inner product between the $m$-th query and $n$-th key in attention score computation as:
\begin{equation}
\begin{aligned}
\label{eq:f_vrope_v1_dot}
    &f_\text{global}(\mathbf{Q}_m,(\mathbf{T}_\text{c2o}^{m})^{\top})^\top f_\text{global}(\mathbf{K}_n,(\mathbf{T}_\text{c2o}^{n})^{-1}) \\
    &=
    \mathbf{Q}_m
    {\bar{\mathbf{R}}}_{d\times d}^{\text{global}}(\mathbf{T}_\text{c2o}^m)
    {\bar{\mathbf{R}}}_{d\times d}^{\text{global}}((\mathbf{T}_\text{c2o}^{n})^{-1})
    \mathbf{K}_n^{\top} \\
    &=
    \mathbf{Q}_m
    {\bar{\mathbf{R}}}_{d\times d}^{\text{global}}\left(\mathbf{T}_\text{c2o}^m(\mathbf{T}_\text{c2o}^{n})^{-1}\right)
    \mathbf{K}_n^{\top} \\
    &=
    \mathbf{Q}_m\left( \mathbf{I}_{d/4 \times d/4}\otimes\left(\mathbf{T}_\text{c2o}^m(\mathbf{T}_\text{c2o}^{n})^{-1}\right)\right) \mathbf{K}_n^{\top}.
\end{aligned}
\end{equation}
As shown in~\cref{eq:f_vrope_v1_dot}, the resulting attention score is conditioned on the relative object pose: $\mathbf{T}_\text{c2o}^m(\mathbf{T}_\text{c2o}^{n})^{-1}$. We illustrate the mechanism of \videoropeone in \cref{fig:videorope}~(a). Despite its effectiveness, it suffers from two major limitations. As shown in \cref{eq:f_vrope_v1_dot} and \cref{fig:videorope}~(a), it simply broadcasts the 3D transformation $\mathbf{T}_\text{c2o}$ to all image tokens associated with an object in each frame. Consequently, it treats the object region as a whole entity rather than establishing pixel-level relationships. Second, \videoropeone degenerates into a frequency-agnostic representation. The channel-wise frequencies in RoPE encode positional information at different scales: high-frequency channels capture local relationships, whereas low-frequency channels model long-range dependencies. In contrast, \videoropeone applies the same transformation uniformly across all feature channels, discarding the multi-frequency spectrum. Both of the above factors constrain the representational capacity of \videoropeone, motivating the development of a pixel-level and frequency-aware relative pose representation. 

\begin{figure*}[!t]
    \centering
    \includegraphics[width=\linewidth]{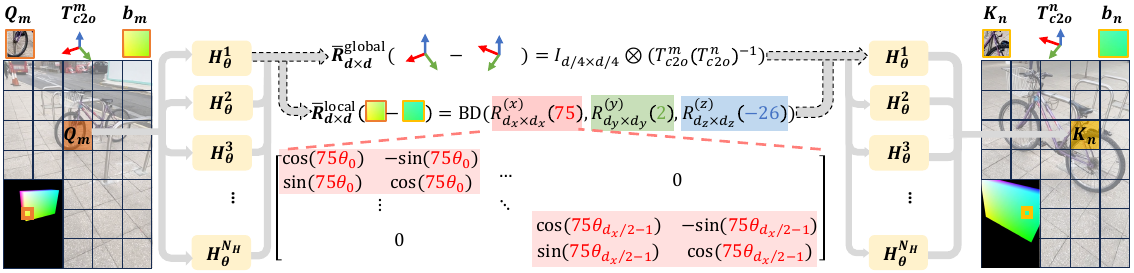}
    \caption{The computation of attention scores in VPRoPE, \ie $f_\text{global/local}(\mathbf{Q}_m,(\mathbf{T}_\text{c2o}^m)^\top/\mathbf{b}_m)^\top f_\text{global/local}(\mathbf{K}_n,(\mathbf{T}_\text{c2o}^n)^{-1}/\mathbf{b}_n)$. For each attention head, $\mathbf{Q}_m$ and $\mathbf{K}_n$ are first transformed by the same head-specific orthogonal matrix $\mathbf{H}_{\theta,h}$, which maps them to a learned orthonormal basis. The transformed attention features are then used in the subsequent inner-product computation. 
    As shown in the middle of the figure, the resulting attention score remains explicitly conditioned on the relative object poses. 
    Meanwhile, each attention head learns a distinct orthonormal basis for interpreting the shared pose-dependent rotations, allowing different heads to capture complementary geometric relationships.}
    \label{fig:dot_product}
\end{figure*}

\myparagraph{\videoropetwo}
The limitations of \videoropeone stem primarily from its use of a global camera-to-object transformation $\mathbf{T}_\text{c2o}$, which fails to capture local geometric variations within an object. In contrast, \cnocs inherently encodes dense, pixel-aligned geometry. Benefiting from this, \videorope-\underline{L} uses the direction angles derived from the per-pixel object-space coordinates within each \underline{l}ocal patch as positional variables.
Specifically, under the identity mapping, $\operatorname{CNOCS}[u,v]$ stores the normalized coordinate $\mathbf{v}=\left[x^{no}_{u,v},y^{no}_{u,v},z^{no}_{u,v}\right]^\top$. The direction angles of $\mathbf{v}$ are given as:
\begin{equation}
\phi^{(a)}
=
\operatorname{atan2}
\left(
\sqrt{
\|\mathbf{v}\|_2^2
-
\left(\mathbf{v}^\top\mathbf{e}_a\right)^2
},
\mathbf{v}^\top\mathbf{e}_a
\right),
\quad a\in\{x,y,z\},
\end{equation}
where $\mathbf{e}_x,\mathbf{e}_y,\mathbf{e}_z$ denote the basis axes of the object coordinate system, and $\phi^{(x)},\phi^{(y)},\phi^{(z)}$ represent the angles between $\mathbf{v}$ and the corresponding axes. Similar to \cref{eq:rope_1d}, we can build the rotary matrix $\mathbf{R}^{(a)}_{d\times d}$ for each dimension as:
\begin{equation}
\begin{aligned}
\label{eq:f_vrope_v2_1d}
&b^{(a)}=\operatorname{Disct}(\phi^{(a)}),\qquad a\in\{x,y,z\}, \\
&\mathbf{R}_{d_a\times d_a}^{(a)}(b^{(a)})= \operatorname{BD}_{i=0}^{d_a/2-1}\!\left(
\begin{bmatrix}
\cos(b^{(a)}\theta_i) & -\sin(b^{(a)}\theta_i)\\
\sin(b^{(a)}\theta_i) & \cos(b^{(a)}\theta_i)
\end{bmatrix}
\right),
\end{aligned}
\end{equation}
where ``$\operatorname{Disct}$'' partitions the angular interval $[ 0,180^\circ]$ into fixed-width bins and maps each continuous angle $\phi^{(a)}$ to its corresponding discrete positional index. Then, we can further derive the formulation of \videoropetwo by combining $\mathbf{R}^{(x)}_{d_x\times d_x},\mathbf{R}^{(y)}_{d_y\times d_y},\mathbf{R}^{(z)}_{d_z\times d_z}$:
\begin{equation}
\begin{aligned}
\label{eq:f_vrope_v2_3d}
&f_\text{local}(\mathbf{x},\mathbf{b}=[b^{(x)},b^{(y)},b^{(z)}])=\bar{\mathbf{R}}_{d\times d}^\text{local}(\mathbf{b})\mathbf{x}^\top\\
&=\operatorname{BD}_{a=x,y,z} \!\left(
\mathbf{R}_{d_a\times d_a}^{(a)}(b^{(a)})
\right)\mathbf{x}^\top.
\end{aligned}
\end{equation}
Thereby, we can calculate the corresponding attention score:
\begin{equation}
\begin{aligned}\label{eq:vrope_v2_dot}
&f_\text{local}(\mathbf{Q}_m,\mathbf{b}_m)^\top f_\text{local}(\mathbf{K}_n,\mathbf{b}_n)\\
&=\mathbf{Q}_{m}\bar{\mathbf{R}}^\text{local}_{d\times d}(\mathbf{b}_{m})^\top\bar{\mathbf{R}}^\text{local}_{d\times d}(\mathbf{b}_{n})\mathbf{K}_{n}^{\top} \\
&=\mathbf{Q}_{m}\bar{\mathbf{R}}^\text{local}_{d\times d}(\mathbf{b}_{n}-\mathbf{b}_{m})\mathbf{K}_{n}^{\top}\\
&=\mathbf{Q}_{m}\operatorname{BD}_{a=x,y,z} \!\left(
\mathbf{R}_{d_a\times d_a}^{(a)}(b_n^{(a)}-b_m^{(a)})
\right)\mathbf{K}_{n}^{\top}.
\end{aligned}
\end{equation}
\cref{fig:videorope}~(b) demonstrates the pixel-aligned property of this encoding mechanism. Compared with \cref{fig:videorope}~(a) and \cref{eq:f_vrope_v1}, \videoropetwo explicitly captures dense, pixel-level geometric relationships while preserving the multi-frequency structure of standard RoPE, thereby enabling richer spatial interactions and better geometric representation capacity.

\myparagraph{Incorporation of Relative Pose Encoding}~Now, the subsequent question is how to incorporate the relative pose encoding into the pretrained T2V model. Our goal is to introduce only a small number of trainable parameters, enabling efficient optimization while largely preserving the motion prior of the original model. 
To this end, \videomodel introduces a set of learnable \textit{head-wise orthogonal transformations} $\mathbf{H}_{\theta}=\left\{\mathbf{H}_{\theta,h}
\in\mathbb{R}^{d\times d}\right\}_{h=1}^{N_{{H}}}$, where $\mathbf{H}_{\theta,h}^{\top}\mathbf{H}_{\theta,h}=\mathbf{I}$ for each attention head $h$. Specifically, $\mathbf{H}_{\theta}$ maps the query and key features into a head-specific orthonormal basis before the pose-dependent rotary transformation, while constraining the transformation to preserve feature norms. 
This allows each head to encode pose relationships in a distinct learned subspace, thereby capturing complementary geometric and motion variations.
To enforce orthogonality without constrained optimization, we parameterize
each transformation as
\begin{equation}\label{eq:ortho}
\mathbf{H}_{\theta,h}
=
\exp\left(
0.5({\mathbf{W}}_{\theta,h}
-
{\mathbf{W}}_{\theta,h}^{\top})
\right),
\end{equation}
where ${\mathbf{W}}_{\theta,h}$ denotes the learnable parameters and $\exp$ is the matrix exponential. 
Therefore, we can reformulate \videoropeone and \videoropetwo as:
\begin{equation}
\begin{aligned}\label{eq:f_vrope_3d_head}
f_\text{global/local}(\mathbf{x},\mathbf{T}_\text{c2o}/\mathbf{b})=\bar{\mathbf{R}}_{d\times d}^\text{global/local}(\mathbf{T}_\text{c2o}/\mathbf{b})\mathbf{H}_{\theta,h}\mathbf{x}^\top.
\end{aligned}
\end{equation}
For better illustration, \cref{fig:dot_product} presents the computation process of the resulting attention score under \videorope.
Taking the example of \videoropetwo, the calculation of the attention score can be written as $\mathbf{Q}_{m} \mathbf{H}_{\theta,h}^\top\bar{\mathbf{R}}^\text{local}_{d\times d}(\mathbf{b}_{n}-\mathbf{b}_{m})\mathbf{H}_{\theta,h}\mathbf{K}_{n}^{\top}$.
Therefore, the resulting attention interaction still depends on the
relative angular distance $\mathbf{b}_{n}-\mathbf{b}_{m}$, while each head learns a
different basis for interpreting the shared pose-dependent rotations.
Finally, we integrate the proposed \videoropetwo with the original RoPE given in \cref{eq:rope_3d}, yielding the final form of our relative pose encoding:
\begin{equation}
\begin{aligned}
\label{eq:f_final}
f_\text{final}(\mathbf{x},\mathbf{p},\mathbf{T}_\text{c2o}/\mathbf{b)}&=f_{\text{RoPE-3D}}(\mathbf{x},\mathbf{p})\\
&+\lambda_\theta \mathbb{I}_{\mathrm{obj}}f_\text{global/local}(\mathbf{x},\mathbf{T}_\text{c2o}/\mathbf{b}),
\end{aligned}
\end{equation}
where $\mathbf{\lambda}_\theta$ is a learnable coefficient initialized to ${0}$. $\mathbb{I}_{\mathrm{obj}}$ is a scalar indicator that equals $1$ if the token is within the object region and $0$ otherwise, ensuring that \videorope is applied only to tokens associated with pose control. Then, the attention score is given as:
{\footnotesize
\begin{equation}
\begin{aligned}
\label{eq:dot_product_final}
&f_{\text{final}}
(
\mathbf{Q}_m,\mathbf{P}_m,
(\mathbf{T}_{\text{c2o}}^m)^\top/\mathbf{b}_m
)^\top f_{\text{final}}
(
\mathbf{K}_n,\mathbf{P}_n,
(\mathbf{T}_{\text{c2o}}^n)^{-1}/\mathbf{b}_n
)
=\\
&\underbrace{
f_{\text{RoPE-3D}}^\top(\mathbf{Q}_m,\mathbf{P}_m)
f_{\text{RoPE-3D}}(\mathbf{K}_n,\mathbf{P}_n)
}_{\text{ordinary RoPE, content}\leftrightarrow\text{content}}
\\
&+\underbrace{
\lambda_\theta
\mathbb{I}_{\mathrm{obj}}^n
(f_{\text{RoPE-3D}}^m)^\top
f_{\text{global/local}}^n
+
\lambda_\theta
\mathbb{I}_{\mathrm{obj}}^m
(f_{\text{global/local}}^m)^\top
f_{\text{RoPE-3D}}^n
}_{\text{content}\leftrightarrow\text{pose}}
\\
&+\underbrace{
\lambda_\theta^2
\mathbb{I}_{\mathrm{obj}}^m
\mathbb{I}_{\mathrm{obj}}^n
(f_{\text{global/local}}^m)^\top
f_{\text{global/local}}^n
}_{\text{pose}\leftrightarrow\text{pose}}.
\end{aligned}
\end{equation}
}
Compared with the ordinary RoPE, the ``$\text{content}\leftrightarrow \text{pose}$'' term calculates the interactions between the original visual content and the pose-conditioned representation, while the last ``$\text{pose}\leftrightarrow \text{pose}$'' term captures the pose similarity. The training objective in this stage is given as:
\begin{equation}
\begin{aligned}
\label{eq:loss_total_stage2}
&\mathcal{L}_{\mathrm{flow,T2V}}=
\mathbb{E}_{\substack{t,(\mathbf{z}_0,c_p,\left\{\text{CNOCS}_i\right\}_{i=1}^F)\sim\mathcal{D}_\text{T2V},\epsilon\sim\mathcal{N}(\mathbf{0},\mathbf{I})}} \\
&\left[
\left\|v_\theta\!\left(\mathbf{z}_t,t,c_p,\left\{\text{CNOCS}_i\right\}_{i=1}^F\right)
-(\epsilon-\mathbf{z}_0)\right\|_2^2
\right], \\
&\mathcal{L}_{\text{Stage-II}}=\mathcal{L}_{\mathrm{flow,T2V}}+\gamma \mathcal{L}_{\mathrm{region}},
\end{aligned}
\end{equation}
where $F$ is the number of frames and $\mathcal{D}_\text{T2V}$ denotes the video subset from \videodata.


\section{Experiments}\label{sec:experiment}
\subsection{Implementation Details}
\videomodel is built upon Wan-T2V-1.3B~\cite{wan2025wan} and initialized with its official pretrained weights. 
During training, we use the Adam optimizer with a learning rate of $1\times10^{-4}$ and a global batch size of 16. 
We implement both $\Phi_\theta^{\text{cnocs}}$ and $\Phi_\theta^{\text{proj}}$ as MLPs, leveraging GELU as activation function.
For the dimensional allocation of $d_x,d_y,d_z$ in \videoropetwo, we adopt the same partitioning strategy as the original RoPE of the pretrained model across the temporal, height, and width dimensions. The bin width of \videoropetwo is set to 1.
For training, the first and second stages are learned for 6K and 12K iterations, respectively. Furthermore, the regional-loss weight $\gamma$ is set to $0.3$ in both stages.

\subsection{Evaluation Details}
\myparagraph{Baselines}~We mainly compare our method with the following models built upon the same foundation model. The first two settings take~\textbf{(1)~2D} and ~\textbf{(2)~3D bounding box} (\textbf{2D/3D BBox}) as control signals, demonstrating the orientation-aware properties of \cnocs. 
Specifically, for the method conditioned on 3D BBox, we directly project the oriented 3D cuboids into the image plane to obtain rendered binary masks. For 2D BBox, we further compute the minimum enclosing rectangles of the projections. 
Furthermore, we leverage the input-level conditioning mechanism in \cref{eq:cnocs_spatial} and optimize the introduced modules on T2I and T2V tasks using the same training configuration as \videomodel. 
To illustrate the effectiveness of the pixel-aligned guidance provided by \cnocs, we also learn a model conditioned on ~\textbf{(3)~implicit pose embedding} (\textbf{IPE}), which has been adopted in several previous studies~\cite{fu20253dtrajmaster,cheng2024learning}.
In particular, the 3D transformation is first flattened and projected to form a 1D vector, which is subsequently appended to the visual tokens and incorporated into the network through the attention mechanism. The newly introduced LoRA modules~\cite{hu2022lora} are optimized during two-stage training. To validate the impact of \videorope, we additionally train a model with~\textbf{(4)~only absolute pose encoding applied }(\textbf{w/o Rel.}). For this setting, we only inject and optimize the modules associated with the absolute pose encoding.

\myparagraph{Construction of \videobench}~To comprehensively evaluate 9D pose control across both T2I and T2V tasks, we introduce \textit{\videobench}. Each test case consists of a text prompt and a 9D pose condition specifying the 3D location, size, and orientation of the objects. 
To simulate realistic spatial configurations and motion patterns, we extract the pose information from real-world images/videos.
For T2I evaluation, we randomly sample images from MS-COCO~\cite{lin2014microsoft}. For T2V generation, the benchmark combines samples from the Objectron test set~\cite{ahmadyan2021objectron}, as well as the videos collected from SpatialVID~\cite {wang2026spatialvid}. Specifically, the datasets without ground-truth pose annotations are processed using the pipeline described in \cref{sec:method_dataset}. Then, we can conveniently convert the obtained 9D poses to the required input representation of each compared model.

Subsequently, we employ an LLM to generate diverse text prompts conditioned on the number of objects. Given the pose condition containing $M$ objects, the LLM is instructed to produce scene descriptions that mention exactly $M$ compatible objects while avoiding spatial or orientation descriptions. We use the following instruction prompt:
\vspace{2pt}

\noindent{\small\itshape
\noindent\textit{Generate [Insert the number of the prompts here] concise prompts describing a scene containing exactly [Insert the number of objects here] objects. Use only object categories with a clearly defined canonical front, such as cars, buses, bicycles, chairs, dogs, or horses. Avoid categories whose front orientation is ambiguous or undefined, such as balls, bowls, bottles, trees, or rocks. Randomly assign each object to be either stationary or moving. For every object, include a compatible action or state description, such as ``a parked car,'' ``a dog running,'' ``a horse standing,'' or ``a cyclist riding.'' Do not describe camera movement. Diversify the environment, object appearance, illumination, and visual style, but do not explicitly specify object locations, sizes, orientations, or motion directions.}
\par}
\vspace{3pt}


Overall, \videobench contains 255 T2I and 326 T2V test cases. For evaluation metrics, we assess the compared methods from the following aspects. 
~\textbf{(1)~Pose accuracy.}~We use SAM 3D to estimate each target object's center $\hat{\mathbf{l}}$ and size $\hat{\mathbf{s}}$ from the generated image and compare them with the input controls ${\mathbf{l}}$ and $\mathbf{s}$ in a common camera frame and at a consistent scene scale.
We define the location error as $\mathrm{Trans.Err}=\|\hat{\mathbf{l}}-\mathbf{l}\|^2_2/\|\mathbf{s}\|^2_2$ and the size error as $\mathrm{Size.Err}=\frac{1}{3}\sum_{k=1}^{3}|\hat{\mathbf{s}}_{k}-\mathbf{s}_{k}|/\mathbf{s}_{k}$, where lower values indicate better accuracy.
For orientation accuracy, we follow previous works~\cite{he2024cameractrl,fu20253dtrajmaster} to report the rotation angle error $\text{Rot.Err}$.
\textbf{(2)~Generation quality.}~Using the images and videos generated by the pretrained model~\cite{wan2025wan} as the reference, we report FID and FVD on \videobench. We also present PickScore~\cite{kirstain2023pick} and HPS-v2.1~\cite{wu2023human} to reflect human preference.
\textbf{(3)~User study.}~Given the lack of dedicated metrics for assessing the coherence of localized object motion, we additionally conduct a user study as a complementary evaluation. We ask participants to assess the pose fidelity and motion coherence. Here, the motion coherence is assessed exclusively on the T2V subset of \videobench. The evaluation criteria are defined as follows.
\begin{itemize}[leftmargin=0.96em]
    \item \textbf{Pose Fidelity $\mathrm{US}_{\mathrm{pos}}$:} Do the objects in the generated image/video follow the target location, size, and orientation? Provide a score between 1 and 10.
    
    \item \textbf{Motion Coherence $\mathrm{US}_{\mathrm{mot}}$:} Does the video exhibit smooth and natural object motion, without rigid transitions, temporal artifacts, or unnatural deformations? Provide a score between 1 and 10.

\end{itemize}
$40$ participants are asked to evaluate anonymized images/videos generated by different methods using the same text prompt, target pose, and seed. The final scores are averaged across all cases.

\begin{table}[t]
\footnotesize
\centering
\caption{Quantitative comparison in terms of 9D pose accuracy and motion coherence. Lower $\text{Trans.Err}$, $\text{Size.Err}$, and $\text{Rot.Err}$ indicate better pose control, while higher $\text{US}_\text{pos}$ and $\text{US}_\text{mot}$ scores demonstrate better pose fidelity and motion coherence. Our method achieves the best overall pose accuracy, with \videoropetwo yielding the most coherent object motion.}
\label{tab:quantitative}
\setlength{\tabcolsep}{2pt}
\resizebox{0.49\textwidth}{!}{
\begin{tabular}{c|l|ccc|cc}
\specialrule{0.7pt}{0pt}{0pt}
\multirow{2}{*}{\textbf{Task}}
& \multirow{2}{*}{\textbf{Method}}
& \multicolumn{3}{c|}{\textbf{9D Pose Accuracy}}
& \multicolumn{2}{c}{\textbf{User Study}} \\
\cline{3-7}
&
& $\text{Trans.Err}$ $\downarrow$
& $\text{Size.Err}$ $\downarrow$
& $\text{Rot.Err}$ $\downarrow$
& $\text{US}_{\mathrm{pos}}$ $\uparrow$
& $\text{US}_{\mathrm{mot}}$ $\uparrow$ \\
\hline
\multirow{4}{*}{T2I}
& 2D BBox
& 0.31
& 0.71
& 1.73
& 4.9
& -- \\
& 3D BBox
& \underline{0.09}
& \underline{0.28}
& 1.10
& \underline{6.5}
& -- \\
& IPE
& 0.23
& 0.67
& \underline{0.58}
& 6.2
& -- \\
& \textbf{Ours}
& \textbf{0.06}
& \textbf{0.19}
& \textbf{0.24}
& \textbf{9.2}
& -- \\
\hline
\multirow{6}{*}{T2V}
& 2D BBox
& 0.42
& 0.78
& 1.92
& 4.6
& 5.3 \\
& 3D BBox
& 0.17
& 0.34
& 1.45
& 6.3
& 5.4 \\
& IPE
& 0.36
& 0.65
& 0.84
& 6.4
& {7.3} \\
& w/o Rel.
& 0.11
& \underline{0.29}
& 0.35
& {8.1}
& 6.2 \\
& \textbf{Ours + \videoropeone}
& \textbf{0.07}
& {0.32}
& \underline{0.31}
& \underline{8.4}
& \underline{7.7} \\
& \textbf{Ours + \videoropetwo}
& \underline{0.08}
& \textbf{0.27}
& \textbf{0.29}
& \textbf{8.5}
& \textbf{8.1} \\
\specialrule{0.7pt}{0pt}{0pt}
\end{tabular}
}
\end{table}

\begin{figure*}[!t]
    \centering \includegraphics[width=\linewidth]{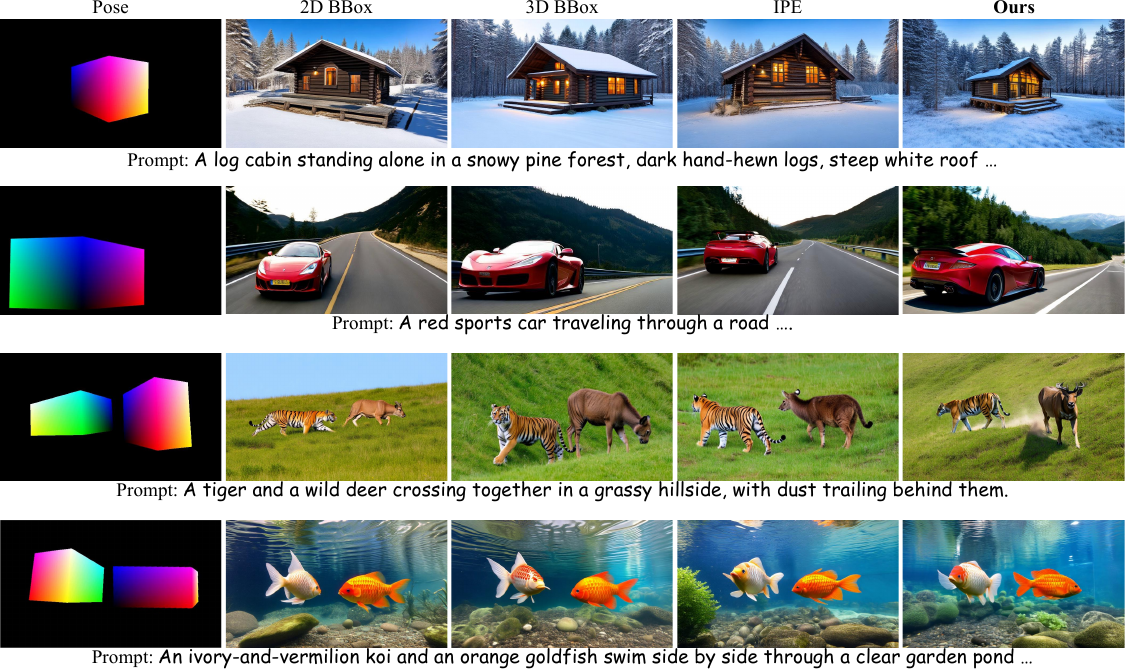}
    \vspace{-5mm}
    \caption{Qualitative comparison of 9D pose control on the T2I task. As illustrated in the figure, models conditioned on 2D or 3D bounding boxes (2D/3D BBox) suffer from orientation ambiguity. Although the implicit pose embedding (IPE) alleviates this issue, its spatially agnostic nature limits the overall accuracy. In contrast, our \videomodel achieves substantially better pose fidelity.}
    \label{fig:t2i_compare}
\end{figure*}

\begin{figure*}[!t]
    \centering
\includegraphics[width=\linewidth]{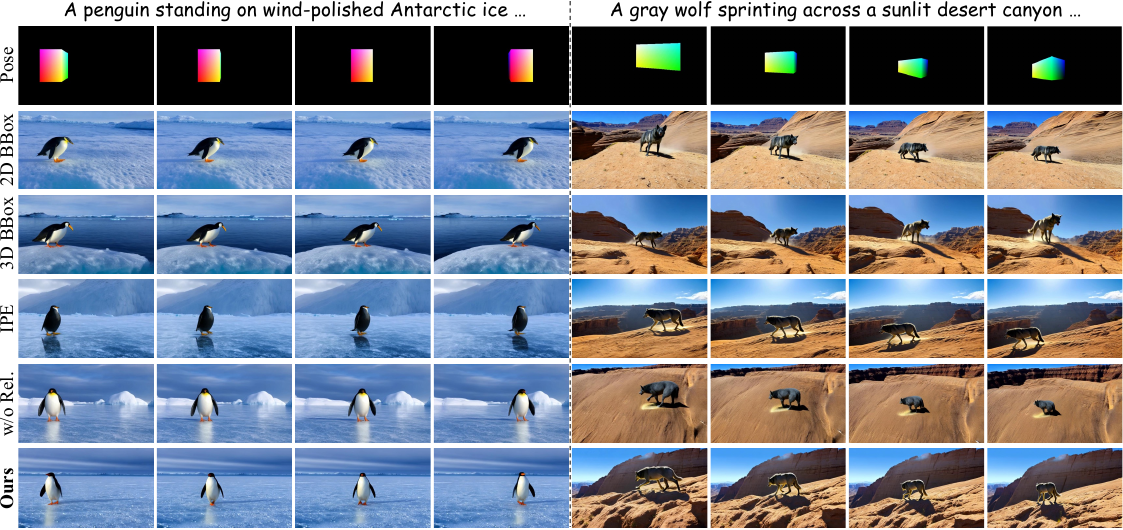}
\vspace{-6mm}
    \caption{Qualitative comparison of 9D pose control on the T2V task. Consistent with the T2I results, most methods exhibit limited pose accuracy, except for CNOCS-based approaches (\ie, w/o Rel., \textbf{Ours}). Moreover, simply conditioning the model on the 2D/3D BBox or absolute pose encoding often results in stiff or unnatural object motion, as evident in the 2nd, 3rd, and 5th rows.}
    \label{fig:t2v_compare}
\end{figure*}

\begin{table*}[!t]
\footnotesize
\centering
\caption{Ablation studies. All variants are implemented with \videoropetwo.}
\label{tab:ablation}
\setlength{\tabcolsep}{7.6pt}
\resizebox{\textwidth}{!}{
\begin{tabular}{c|l|ccc|cccc}
\specialrule{0.7pt}{0pt}{0pt}
\multirow{2}{*}{\textbf{Task}}
& \multirow{2}{*}{\textbf{Method}}
& \multicolumn{3}{c|}{\textbf{9D Pose Accuracy}}
& \multicolumn{4}{c}{\textbf{Visual Quality}} \\
\cline{3-9}
&
& $\text{Trans.Err}$ $\downarrow$
& $\text{Size.Err}$ $\downarrow$
& $\text{Rot.Err}$ $\downarrow$
& $\mathrm{FID}$ $\downarrow$
& $\mathrm{FVD}$ $\downarrow$
& $\mathrm{PickScore}$ $\uparrow$
& $\mathrm{HPSv2}$ $\uparrow$ \\
\hline
\multirow{2}{*}{T2I}
& w/o $L_{\mathrm{region}}$
& \textbf{0.05}
& 0.26
& 0.28
& \textbf{28.1}
& --
& {22.15}
& {0.269} \\
& \textbf{Ours}
& {0.06}
& \textbf{0.19}
& \textbf{0.24}
& {29.7}
& --
& \textbf{22.37}
& \textbf{0.276} \\
\hline
\multirow{3}{*}{T2V}
& w/o $L_{\mathrm{region}}$
& 0.13
& 0.29
& 0.38
& \textbf{30.8}
& \textbf{278.4}
& \textbf{22.26}
& {0.271} \\
& w/ Head-Shared Transformation
& 0.14
& 0.32
& 0.36
& 31.4
& 295.6
& 22.20
& \underline{0.273} \\
& w/o Orthogonality
& \underline{0.09}
& \textbf{0.24}
& \underline{0.34}
& 37.2
& 584.7
& 21.96
& 0.264 \\
& \textbf{Ours}
& \textbf{0.08}
& \underline{0.27}
& \textbf{0.29}
& \underline{31.7}
& \underline{297.9}
& \underline{22.24}
& \textbf{0.275} \\
\specialrule{0.7pt}{0pt}{0pt}
\end{tabular}
}
\end{table*}

\subsection{Qualitative and Quantitative Results}
In this section, we evaluate the compared methods on the proposed \videobench. \cref{fig:t2i_compare} presents qualitative comparisons. Among the compared methods, the model conditioned on 2D/3D BBox can control only the coarse location and size of the object, without explicitly modulating its orientation, resulting in ambiguous outcomes. On the other hand, the model based on IPE struggles to achieve highly precise pose control, as its spatially agnostic nature limits overall accuracy. For example, the car in the 2nd row of \cref{fig:t2i_compare} is placed in the wrong location. In contrast, our \videomodel achieves substantially better pose fidelity. These observations are also evident in the T2V results shown in \cref{fig:t2v_compare}, where most compared methods exhibit poor pose fidelity, except for CNOCS-based approaches (\ie, w/o Rel., \textbf{Ours}). However, since the w/o Rel. setting fails to capture the geometric cues governing pose transitions, it tends to generate unnatural or overly rigid object motion. A similar issue arises in methods based on 2D/3D BBox settings for analogous reasons. As shown in the 2nd, 3rd, and 5th rows of \cref{fig:t2v_compare}, the penguin’s feet and the wolf’s limbs are extremely stiff, resulting in motion that visually resembles rigid-body transformation. In contrast, our method produces accurate object poses with more coherent motion. 

\begin{figure*}[!t]
    \centering
\includegraphics[width=0.99\linewidth]{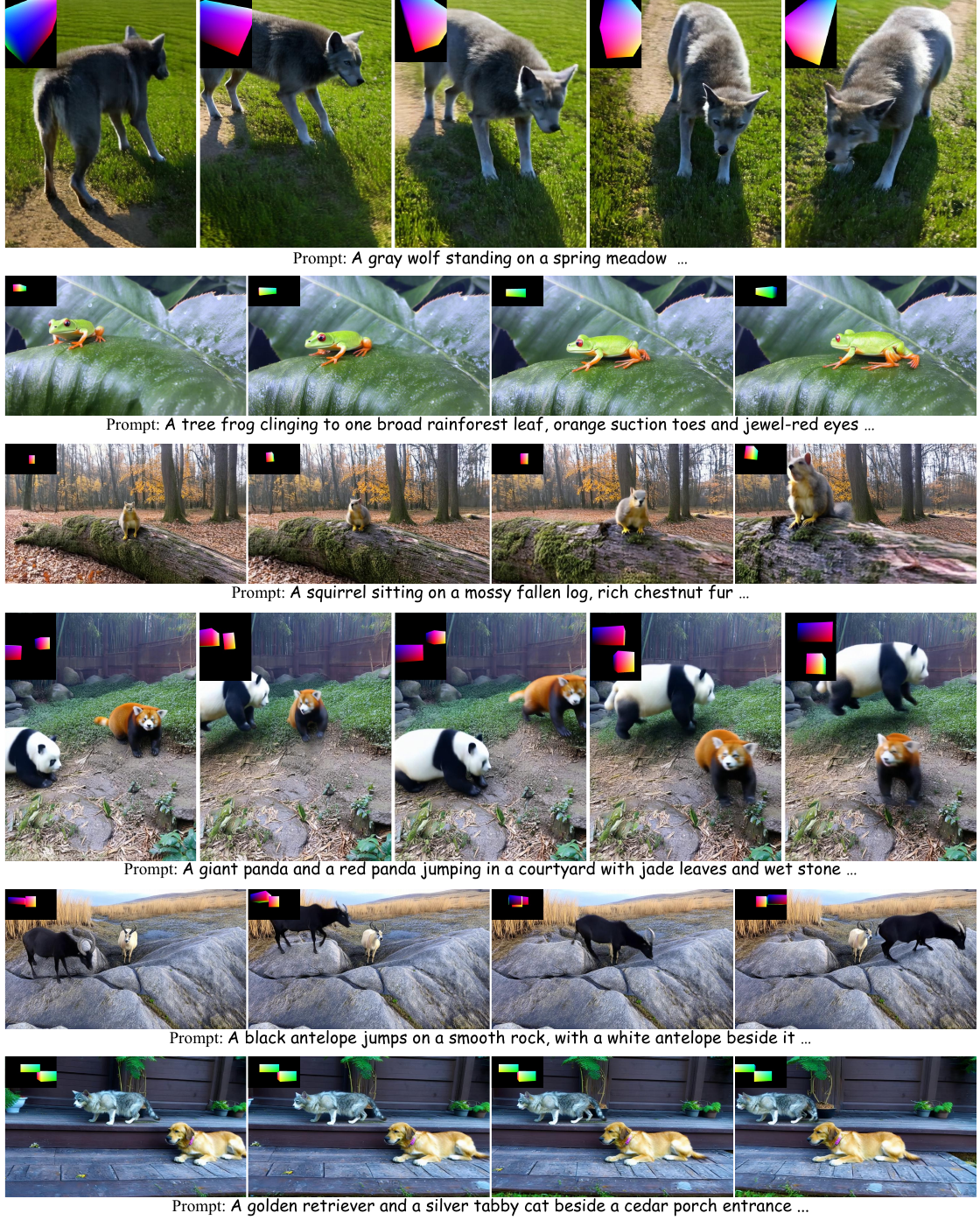}
    \caption{More qualitative results on the T2V task. Our method consistently achieves high pose fidelity across all examples.
    }
    \label{fig:t2v}
\end{figure*}

More T2V results are shown in \cref{fig:t2v}. Several cases in \cref{fig:t2v} also illustrate plausible visual effects of the generated results. Notably, the large-scale motions shown in the 3rd and 4th rows produce realistic motion blur, while the 5th row demonstrates that object appearances remain consistent across mutual occlusions.

Additionally, \cref{tab:quantitative} reports the quantitative results on \videobench. Due to the lack of explicit 3D cues, the 2D BBox baseline performs poorly across all pose metrics. Although 3D BBox partially improves location and size control ($\text{Trans.Err}, \text{Size.Err}$), its coarse geometric representation remains insufficient for accurately characterizing object orientation ($\text{Rot.Err}$). In contrast, our method achieves superior pose accuracy across most dimensions. Compared with the baselines that merely extend frame-wise control signals along the temporal dimension, \videorope conditions attention on relative poses, enabling the model to establish geometry-aware correspondences across frames and produce more coherent object motion, as indicated by $\text{US}_{\text{mot}}$. In particular, \videoropetwo further captures pixel-level geometric correspondences, achieving a higher $\text{US}_{\text{mot}}$ score than \videoropeone.

\subsection{Ablation Studies}
To further evaluate the effectiveness of our design choices, we conduct ablation studies on the regional loss, head-wise transformations, and orthogonal parameterization. Specifically, we compare the full model with the following variants. \textbf{(1) w/o $L_{\mathrm{region}}$} removes the regional loss, allowing us to evaluate the importance of region-aware supervision. \textbf{(2) w/ Head-Shared Transformation} replaces the head-wise transformations with a single orthogonal transformation shared across all attention heads, allowing us to assess the benefit of head-specific geometric modeling. \textbf{(3) w/o Orthogonality} removes the orthogonality constraint in \cref{eq:ortho} and allows each head-wise transformation $\mathbf{H}_{\theta,h}$ to be optimized freely, thereby examining the effect of the orthogonal parameterization. All ablation experiments are conducted with \videoropetwo. 
As shown in \cref{tab:ablation}, removing $L_{\mathrm{region}}$ weakens the supervision on object regions and consequently reduces the overall pose accuracy. Similarly, sharing a single orthogonal transformation across all attention heads limits the representational capacity, also leading to suboptimal pose accuracy. Meanwhile, removing the orthogonality constraint results in degraded generation quality. In contrast, our full model achieves a favorable balance between accuracy and quality.

\subsection{Joint Control of Object and Camera Motions}
\begin{figure*}[!t]
    \centering
\includegraphics[width=\linewidth]{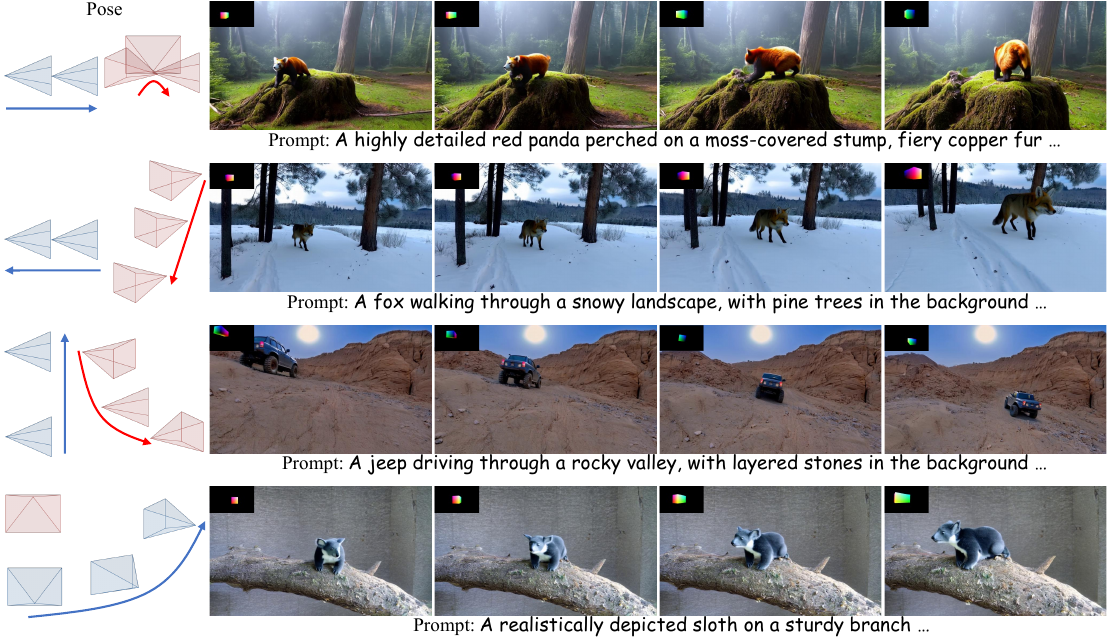}
    \caption{Qualitative results of joint object and camera control. The blue and red trajectories denote camera and object motion, respectively. The generated background motion faithfully follows the specified camera trajectory, while the object poses remain consistent with the input \cnocs sequence throughout the video. For instance, in the 2nd row, the camera slowly recedes while the fox approaches more rapidly, causing the trees to appear smaller and the fox progressively larger.}
    \label{fig:t2v_cam}
\end{figure*}

We integrate \videomodel with an independently trained camera-control module and present the generated results in \cref{fig:t2v_cam}. In particular, the module follows the architecture of UCPE~\cite{zhang2026unified} and is initialized from its weights pretrained on datasets with camera-motion annotations. As shown in \cref{fig:t2v_cam}, the background dynamics consistently follow the specified camera trajectory, while the object motion remains faithful to the poses defined by the \cnocs sequence. For example, in the second row, the camera gradually pulls back, causing the background trees to appear progressively smaller. Meanwhile, the fox moves toward the camera faster than the camera recedes and therefore occupies an increasingly larger region of the frame. 
These examples demonstrate that our method effectively decouples object and camera motion while preserving coherent perspective changes and temporal consistency.

\section{Conclusion and Limitations}
This work introduces \videomodel, which extends our previous conference version, \imagemodel, in several key aspects. First, \videomodel provides a unified framework for 9D pose control across both T2I and T2V tasks, allowing users to flexibly specify object motions in 3D space. To provide rich supervision for learning generalizable pose control, we introduce \videodata, which augments \imagedata with additional images and videos accompanied by comprehensive 9D pose annotations. By training on this dataset, \videomodel combines absolute and relative pose encoding to decouple frame-wise spatial grounding from inter-frame motion modeling. A two-stage training strategy is further devised to learn these two mechanisms separately. The decoupled pose encoding strategy and training scheme enable accurate pose control while producing coherent and natural object dynamics. In addition, \videomodel can be composed with a standalone camera-control module to support both independent and joint control over object and camera motion. Extensive experiments on our \videobench benchmark demonstrate that \videomodel consistently outperforms the baselines in terms of pose fidelity and motion coherence.

\myparagraph{Limitations and Future Work} The support of \videomodel for image-to-video (I2V) generation remains constrained by the image-conditioning capability of the underlying pretrained model. In future work, we plan to investigate more general-purpose generative foundation models and develop a pose-control framework that supports a broader spectrum of conditioning modalities.

\bibliographystyle{IEEEtran}
\bibliography{references}

\vfill

\end{document}